\documentclass{article}

\usepackage{wrapfig}
 \usepackage[final]{nips_2017}

\usepackage[utf8]{inputenc} 
\usepackage[T1]{fontenc}    
\usepackage{hyperref}       
\usepackage{url}            
\usepackage{booktabs}       
\usepackage{amsfonts}       
\usepackage{nicefrac}       
\usepackage{microtype}      
\usepackage{multirow}
\usepackage{color}
\usepackage{enumerate}
\usepackage{amsmath,amssymb} 
\usepackage{pgfplots}
\pgfplotsset{compat=newest}
\usepackage{pdflscape}
\usepackage{graphicx}
\usepackage[export]{adjustbox}
\usepackage{algorithm}
\usepackage{mathtools}
\usepackage{algorithmic}
\usepackage{wrapfig,booktabs}

\usepackage{array}

\usepackage{wrapfig}

\title{Maximal adversarial perturbations for obfuscation: Hiding certain attributes while preserving rest}

\author{\parbox{16cm}{\centering
    {Indu Ilanchezian$^2$, Praneeth Vepakomma$^1$, Abhishek Singh$^1$, Otkrist Gupta$^1$, G.N.S.Prasanna$^2$, Ramesh Raskar$^1$}\\
    {\normalsize
    $^1$ Massachusetts Institute of Technology, Cambridge, U.S.A\\
    $^2$ Indian Institute of Information Technology, Bangalore, India}}
}

\begin{document}

\maketitle

\begin{abstract}

In this paper we investigate the usage of adversarial perturbations for the purpose of privacy from human perception and model (machine) based detection. We employ adversarial perturbations for obfuscating certain variables in raw data while preserving the rest. Current adversarial perturbation methods are used for data poisoning with \textit{minimal} perturbations of the raw data such that the machine learning model's performance is adversely impacted while the human vision cannot perceive the difference in the poisoned dataset due to minimal nature of perturbations. We instead apply relatively \textit{maximal} perturbations of raw data to conditionally damage model's classification of one attribute while preserving the model performance over another attribute. In addition, the maximal nature of perturbation helps adversely impact human perception in classifying hidden attribute apart from impacting model performance. We validate our result qualitatively by showing the obfuscated dataset and quantitatively by showing the inability of models trained on clean data to predict the hidden attribute from the perturbed dataset while being able to predict the rest of attributes.
\end{abstract}

\section{Introduction}
In this paper we investigate the usage of adversarial perturbations for privacy from both human and model (machine) based classification of hidden attributes from the perturbed data. With the advent of distributed learning, several methods such as federated learning and split learning have become prominent for distributed deep learning. At the same time, privacy preserving machine learning is a very active area of research. Adversarial approaches of minimally perturbing the data to fool the model performance while fooling human perception in detecting such a perturbation have become popular. We investigate whether an advesarial perturbation can be used to achieve the following goals: 

\begin{enumerate}
    \item Damage the model performance for predicting a chosen sensitive attribute while keeping the perfomance of predicting another attribute intact.
    \item Obfuscate with maximal perturbation to make it difficult for human to detect the hidden attribute.

\end{enumerate}
Such perturbations are necessitated in sectors like finance \citep{bateni2018fair,srinivasan2019generating,chen2018interpretable,chen2018fair}, healthcare \citep{vepakomma2018split,chang2018distributed}, government, retail \citep{zhao2017men, yao2017beyond} and hiring \citep{kay2015unequal,Harrison2018Bias} due to privacy, fairness, ethical and regulatory issues. \par
We validate our results qualitatively by presenting the perturbed datasets and quantitatively by showing the inability of models trained on clean data to predict the hidden attribute from the perturbed dataset while being able to predict the rest of attributes. We consider these to be useful intermediate experiments and results towards the goal of using adversarial methods for generating perturbations such that when a model is trained from the perturbed data for predicting the hidden attribute, the model performance is under control. We also would like to mention that these approaches can motivate a theoretical study of privacy guarantees of adversarial approaches under worst-case settings. 

\subsection{Contributions and method}

\begin{wrapfigure}{r}{0.5\textwidth}\label{Appp}
  \begin{center}
    \includegraphics[width=0.40\textwidth]{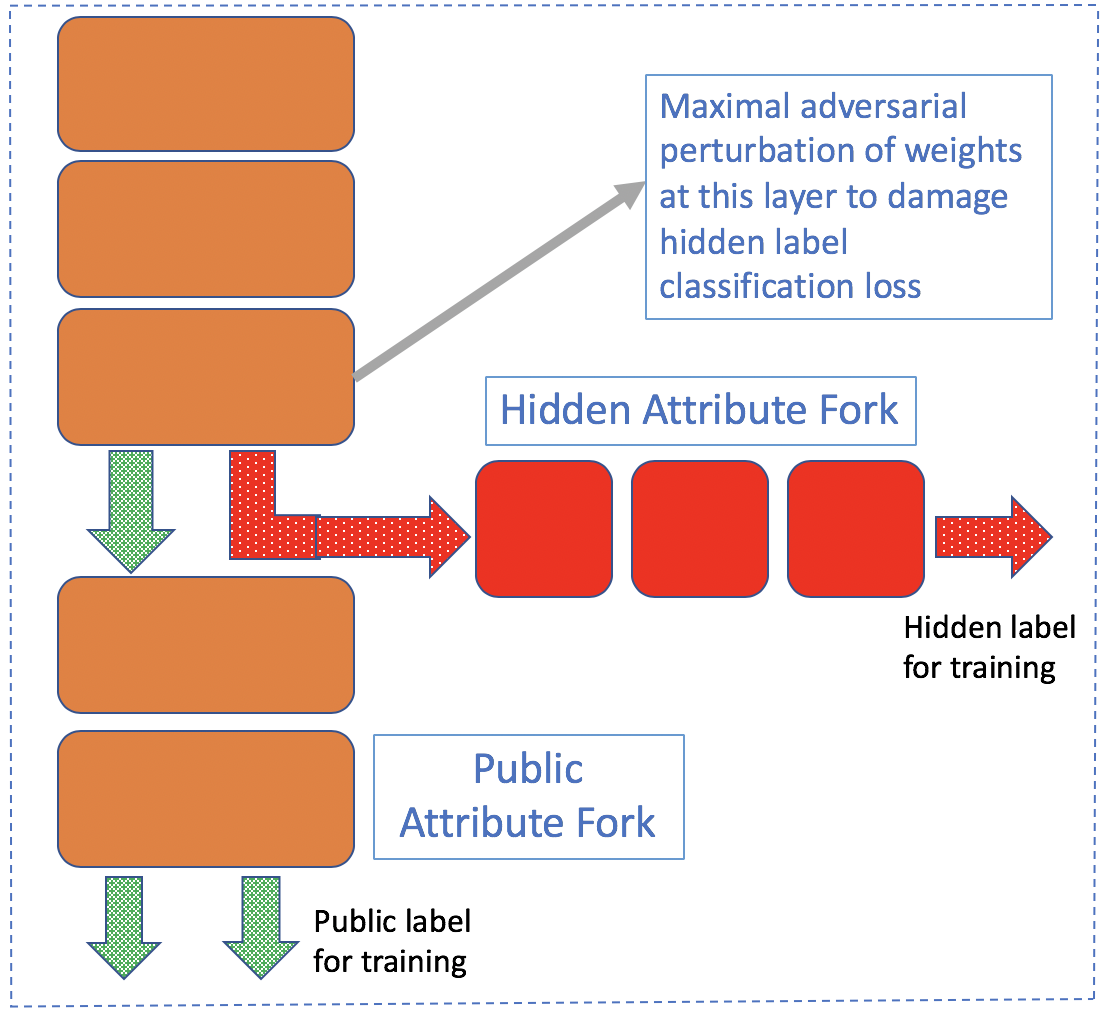}
  \end{center}
  \caption{The architecture of the model used is shown above, with two forks, one to preserve the hidden label attribute and the other to model the public label attribute. Adversarial perturbations are added at the layer preceding the hidden attribute fork.}
\end{wrapfigure}

We employ adversarial perturbation methods with a larger $\epsilon$-ball of perturbations to generate images that are pretty hard for humans to classify with respect to the hidden attribute. Typically adversarial perturbations have only been used with small $\epsilon$-ball perturbations to the best of our knowledge in rest of the works. However, the goal of this work is to obfuscate the data to hide certain latent information from the both model as well as humans, therefore, we conduct our experiments over a broad spectrum of the epsilon values which encompass the output range which fool machine as well as human. In this study we consider only images but our technique is generic in design and applicable wherever adversarial perturbations can be performed successfully. To generate adversarial perturbation, we use VGG-16~\cite{simonyan2014very} model with pretrained layers on ImageNet~\cite{imagenet}. We employ an architecture as shown in Figure \ref{Appp} where a fork is created after three blocks of the VGG-16 architecture where each block consists of two convolution layers and one pooling layer. The hidden attribute fork consists of few layers of DNN for local computation. The rest of the network after the red fork is used to predict the label attribute that is supposed to be preserved. We then train the network. Upon training, we then employ adversarial methods of fast gradient sign method (FGSM) and projected gradient based perturbation (PG), to perturb the layer preceding the hidden attribute fork (shown by grey arrow). We although choose a higher $\epsilon$-ball of possible perturbations in order to generate the perturbation of this layer with respect to loss function corresponding to only the hidden label attribute. We show detailed results with regards to the quality of our results in the experiments section. In addition we weight the two loss functions with $\alpha_1,\alpha_2$.
\subsection{Related work}
\begin{figure}[!htbp]\label{Appp2}
  \begin{center}
    \includegraphics[width=0.80\textwidth]{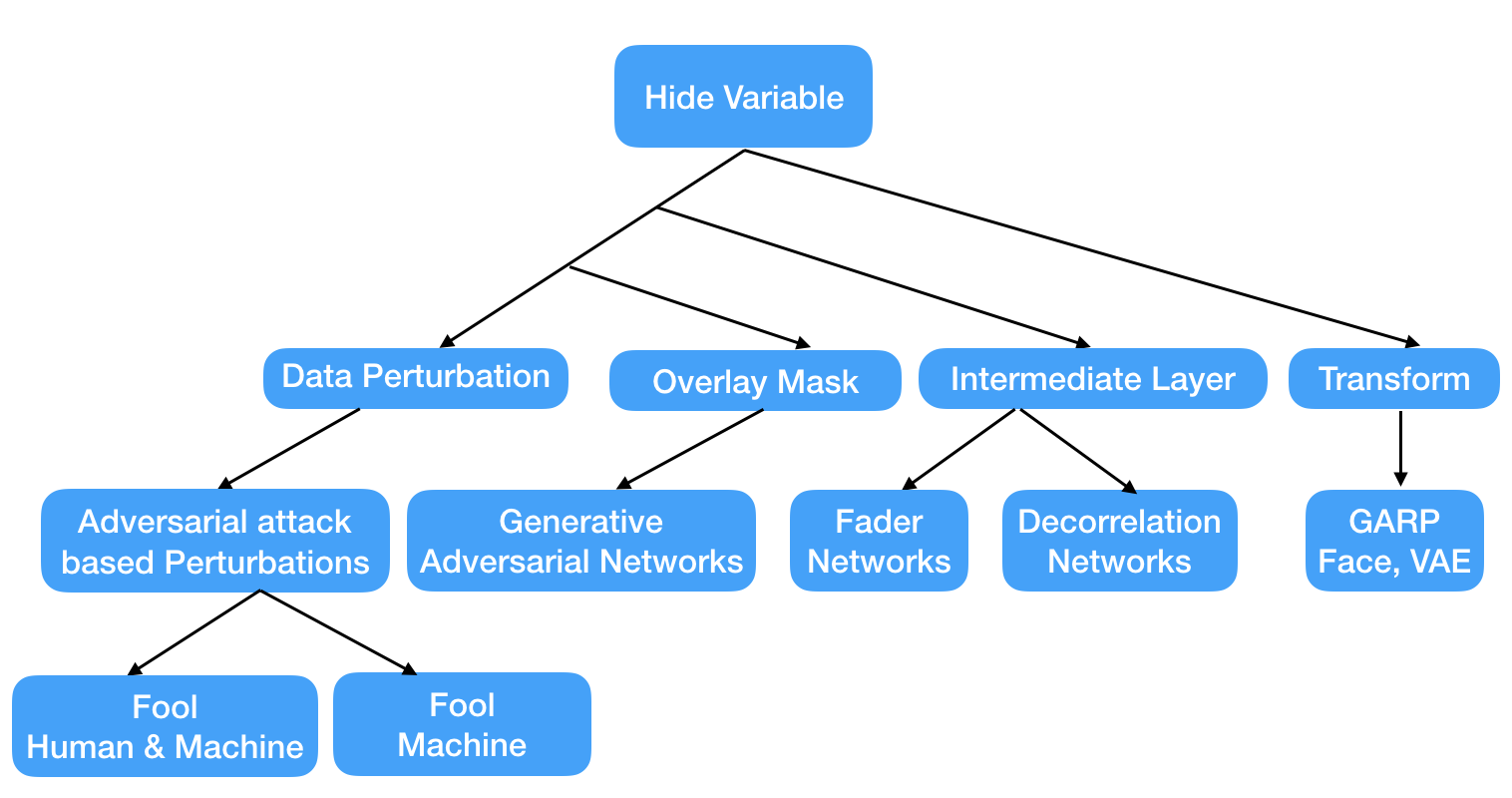}
  \end{center}
\caption{We share the landscape of deep learning based approaches for hiding certain attributes in data. We categorize it broadly into 4 approaches that include a.) perturbation of raw data, b.) overlaying mask on raw data to hide certain parts of image, c.) modifying the output of intermediate (encoded) representations, d.) transforming the entire (or partial) natural image into another natural image. We share 6 example methods within these categories. Our approach belongs to the category of `adversarial attack based perturbations'. As a sub note, all the above approaches can be further categorized into sub approaches that fool humans and/or machines.}
\end{figure}

In Figure \ref{Appp2} we share the landscape of deep learning based approaches for hiding certain attributes in data. We categorize it broadly into 4 approaches that include a.) perturbation of raw data which includes our approach, b.) overlaying mask on raw data to hide certain parts of image, \cite{relbalanceDist}, c.) modifying the output of intermediate (encoded) representations \citep{lample2017fader,vepakomma2018split,vepakomma2019reducing}, d.) transforming the entire (or partial) natural image into another natural image \citep{du2014garp,relchen2019distributed,relwu2018privacy}. We share 6 example methods within these categories. Our approach belongs to the category of `adversarial attack based perturbations'. As a sub note, all the above approaches can be further categorized into sub approaches that fool humans and/or machines while our intermediate work focuses on both.

\section{Results}

We detail our experimental setup and results in this section. A condensed version of the architecture in our setup is shown in figure \ref{Appp}. We perform experiments where VGG16 is the initially trained model that is adversarially perturbed with large $\epsilon$-ball of perturbations. We then predict over the perturbed data with a clean model trained on unperturbed data. Three clean models were trained with architectures of VGG16 and VGG19. The two methods of adversarial perturbations used with large choice of $\epsilon$-ball were fast gradient sign method (FGSM), \cite{goodfellow2014explaining} and projected gradient adversarial method (PGD), \cite{athalye2017synthesizing}. The dataset used was UTKFace which is a large-scale face dataset with long age span. The dataset consists of over 20,000 face images with annotations of age, gender, and ethnicity. The images cover large variation in pose, facial expression, illumination, occlusion, resolution, etc. This dataset could be used on a variety of tasks, e.g., face detection, age estimation, age progression/regression and landmark localization. In Table \ref{tab1}, we show results of the different methods of PGD abd FGSM employed in the large $\epsilon$-ballsetting with different weights $\alpha_1,\alpha_2$ for the weighted loss and the accuracy of predicting on perturbeddata generated by our approach using clean models of VGG16 and VGG19 trained on unperturbed data. We note that $82.92\%$ of race predictions by the clean model after our perturbation belong to the majority race class.  This shows that our method is successfully able to increase the no-information rate in our predictions of the hidden label attribute of race while preserving gender accuracy. $39.2\%$ of ground truth of race belong to the same class as well. Therefore we reach the required level of obfuscation.

\begin{figure}[!htb]
\centering
\includegraphics[width=9cm]{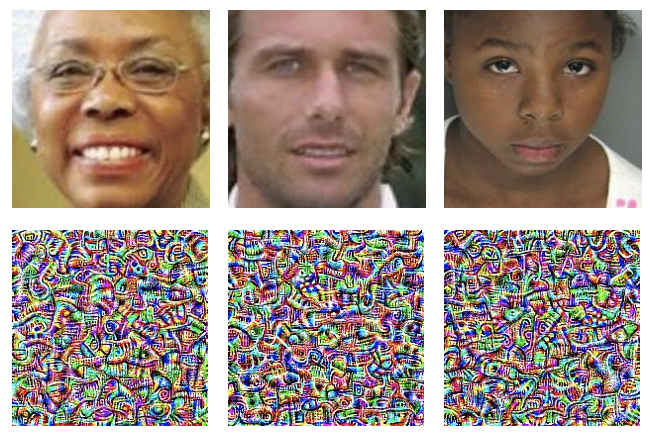}
\caption{Original and perturbed images obtained by projected gradient descent based adversarial perturbation shows a much greater quality of perturbed images. The $\epsilon$ used here is 0.2 and the data belongs to the UTK data set. Race is the attribute preserved and gender is the public attribute. The table 1 shows the corresponding results where the race accuracy is brought down to $0$ as ideally desired while the gender accuracy is relatively preserved.}
\end{figure}

\begin{figure}[!htb]
\centering
\includegraphics[width=9cm]{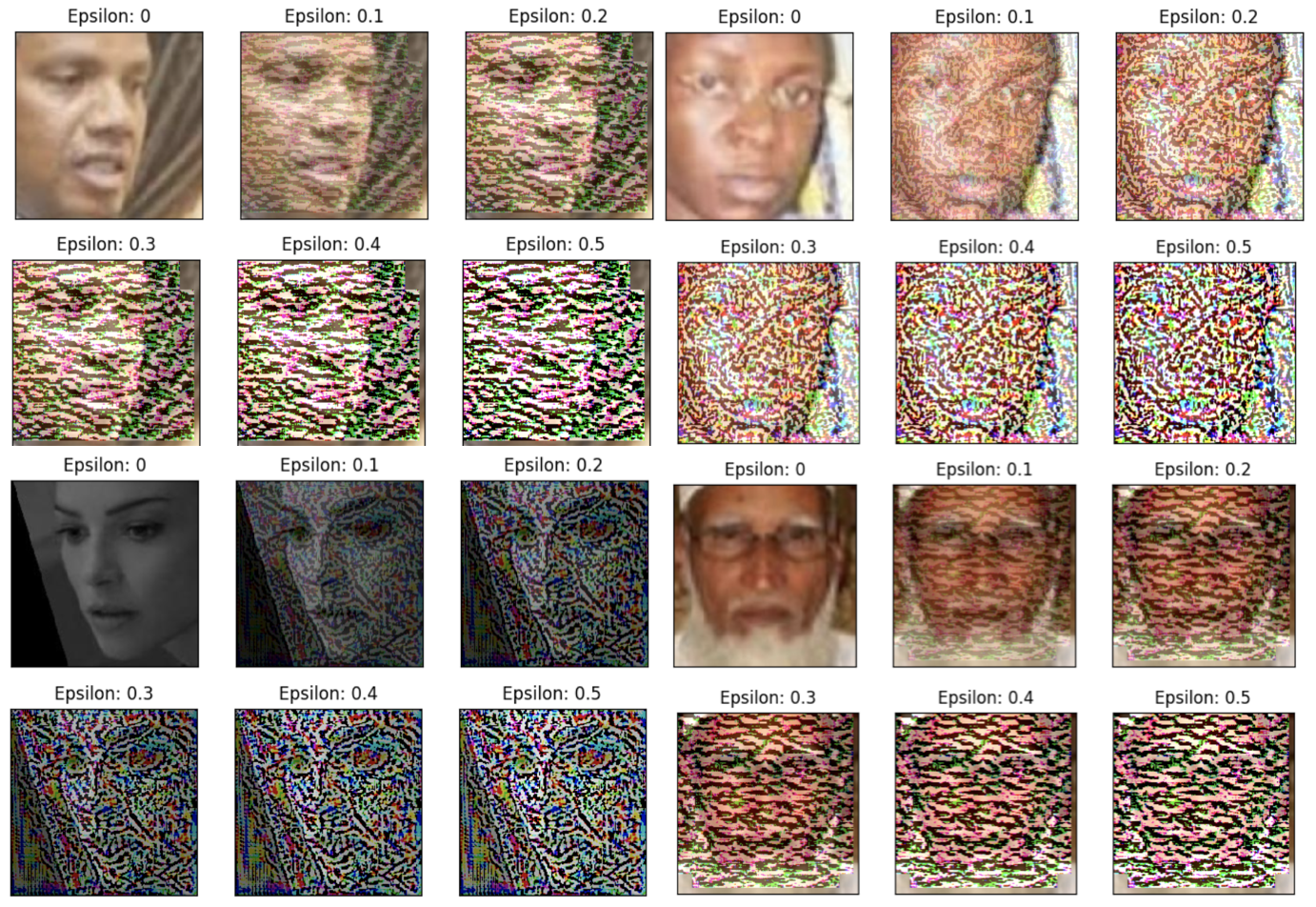}
\caption{Original and perturbed images obtained by fast gradient sign method based adversarial perturbation. The data belongs to the UTK data set. The $\epsilon$'s used here include 0.3, 0.4 and 0.5. Race is the attribute preserved and gender is the public attribute. The table 1 shows the corresponding results where the race accuracy is brought down to $0.21$ while the gender accuracy is relatively preserved at $0.66$. This was a good method to investigate although the projected gradient descent based method has better qualitative and quantitative performance.}
\end{figure}

\begin{table}[]\label{tab1}
\begin{center}
\begin{tabular}{|l|l|l|l|l|l|l|}

\hline
$\alpha_1$ & $\alpha_2$ & Epsilon & Race Accuracy & Gender Accuracy & Method & Clean Model \\ \hline
1          & 1          & 0.2     & 0\%           & 72.9\%          & PGD    & VGG16       \\ \hline
1          & 1          & 0.2     & 30\%          & 54\%            & PGD    & VGG19       \\ \hline
1          & 1.00E-05   & 0.3     & 40\%          & 65\%            & FGSM   & VGG16       \\ \hline
1          & 1.00E-05   & 0.4     & 40\%          & 58\%            & FGSM   & VGG16       \\ \hline
1          & 1.00E-05   & 0.5     & 40\%          & 54\%            & FGSM   & VGG16       \\ \hline
\end{tabular}
\caption{We show results of the different methods of PGD and FGSM employed in the large $\epsilon$-ball setting with different weights $\alpha_1,\alpha_2$ for the weighted loss and the accuracy of predicting on perturbed data generated by our approach using clean models of VGG16 and VGG19 trained on unperturbed data. The baseline performance on clean data prior to perturbation is $87\%$ for race accuracy and $97\%$ for gender accuracy. We note that $82.92\%$ of race predictions by the clean model after our perturbation belong to the majority race class.  This shows that our method is successfully able to increase the no-information rate in our predictions of the hidden label attribute of race while preserving gender accuracy. $39.2\%$ of ground truth of race belong to the same class as well. Therefore we reach the required obfuscation target.}
\end{center}
\end{table}
\section{Conclusion and future work}
We investigate large $\epsilon$-ball perturbations obtained via adversarial methods for obfuscating one label attribute while preserving rest. We show better performance in fooling human with projected gradient descent based approach and better utility in preserving accuracy of public label attributes with fast gradient sign method based prediction. For future work, we aim to enhance this approach with information theoretic and other statistical dependency minimizing loss functions like distance correlation, Hilbert Schmidt Independence Criterion and Kernel Target Alignment. We note that $82.92\%$ of race predictions by the clean model after our perturbation belong to the majority race class.  This shows that our method is successfully able to increase the no-information rate in our predictions of the hidden label attribute of race while preserving gender accuracy. $39.2\%$ of ground truth of race belong to the same class as well. Therefore we reach the required baseline. Therefore the other goal would be to raise the performance of predicting the public label attribute as we reach the required obfuscation performance on hidden label attribute.

\bibliographystyle{authordate1} 
\bibliography{expert_matcher}

\end{document}